\documentclass[
twocolumn, 
hf, 
]{ceurart}

\usepackage{listings}
\begin{document}

\copyrightyear{2025}
\copyrightclause{Copyright for this paper by its authors.
  Use permitted under Creative Commons License Attribution 4.0
  International (CC BY 4.0).}

\conference{\href{https://hexed-workshop.github.io}{HEXED'25: 2nd Human-Centric eXplainable AI in Education Workshop}, 20 July, 2025, Palermo, Italy}


\title{\textcolor{red}{Accepted to appear in the workshop proceedings for the HEXED'25 workshop in the 26th International Conference on Artificial Intelligence in Education 2025} \newline \newline Explainable Collaborative Problem Solving Diagnosis with BERT using SHAP and its Implications for Teacher Adoption}


\author[1]{Kester Wong}[%
orcid=0000-0002-8689-6869,
email=yew.wong.21@ucl.ac.uk
]
\cormark[1] 
\address[1]{UCL Knowledge Lab, Institute of Education, University College London, UK}

\author[2]{Sahan Bulathwela}[%
orcid=0000-0002-5878-2143,
email=m.bulathwela@ucl.ac.uk
]
\address[2]{UCL Centre for Artificial Intelligence, Department of Computer Science, University College London, UK}

\author[1]{Mutlu Cukurova}[%
orcid=0000-0001-5843-4854,
email=m.cukurova@ucl.ac.uk
]

\cortext[1]{Corresponding author.}

\begin{abstract}
  The use of Bidirectional Encoder Representations from Transformers (BERT) model and its variants for classifying collaborative problem solving (CPS) has been extensively explored within the AI in Education community. However, limited attention has been given to understanding how individual tokenised words in the dataset contribute to the model’s classification decisions. Enhancing the explainability of BERT-based CPS diagnostics is essential to better inform end users such as teachers, thereby fostering greater trust and facilitating wider adoption in education. This study undertook a preliminary step towards model transparency and explainability by using SHapley Additive exPlanations (SHAP) to examine how different tokenised words in transcription data contributed to a BERT model's classification of CPS processes. The findings suggested that well-performing classifications did not necessarily equate to a reasonable explanation for the classification decisions. Particular tokenised words were used frequently to affect classifications. The analysis also identified a spurious word, which contributed positively to the classification but was not semantically meaningful to the class. While such model transparency is unlikely to be useful to an end user to improve their practice, it can help them not to overrely on LLM diagnostics and ignore their human expertise. We conclude the workshop paper by noting that the extent to which the model appropriately uses the tokens for its classification is associated with the number of classes involved. It calls for an investigation into the exploration of ensemble model architectures and the involvement of human-AI complementarity for CPS diagnosis, since considerable human reasoning is still required for fine-grained discrimination of CPS subskills.
\end{abstract}


\maketitle

\section{Introduction}
Collaborative problem solving (CPS) is a multilevel process involving the perspectives of both individuals and groups, and should take into account the cognitive domain knowledge that is activated during CPS \cite{luckin_2017Solved}. As such, different conceptual frameworks that predominantly focus on social and cognitive constructs of CPS, with varying granularity and complexities of facets or skills, have been used in various studies to diagnose CPS competencies or processes in student interactions \cite{hesse_2015Framework, cukurova_2016analysis, taylor_2024Quantifying}. Natural language processing (NLP) techniques and different modelling approaches can be leveraged to enable automated diagnosis of these competencies or processes. Automated diagnosis presents an opportunity to enhance research methodology by improving the efficiency and consistency of qualitative coding, while also enabling automated support for students' CPS development in computer-mediated environments \cite{dmello_2024Improving}.

Several existing studies have investigated the use of Bidirectional Encoder Representations from Transformers (BERT) and its variant models for CPS diagnosis. For example, \citet{pugh_2021Say} implemented a model (i.e., BERT-seq) that used the previous utterance and the subsequent utterance to determine the final classification of a given utterance. This approach enabled the model to capture contextual information, improving the area under the receiver operating characteristic curve (AUROC) scores for some classes. A recent study by \citet{zhao_2024automated} investigated the use of two pre-trained BERT models to determine aspects of embodied teamwork when resolving issues in clinical simulation tasks; Conversational BERT (ConvBERT) which uses convolution layers to address local dependencies in natural language \cite{NEURIPS2020_96da2f59}, and CliBERT which is trained on clinical text \cite{alsentzer-etal-2019-publicly}. However, these studies only involved the use of unimodal BERT-variant models with transcribed conversations. On the other hand, \citet{wong_2025Rethinking} explored the use of a multimodal BERT model (i.e., AudiBERT\footnote{\url{https://github.com/keswong/AIED25-AudiBERT}}) that used both textual and acoustic features of speech data. The study found that while there were observable marginal improvements in the diagnosis of specific CPS subskills in the social-cognitive dimension, the BERT model demonstrated higher precision, accuracy, and recall for classifying CPS affective states. Despite the recognised value of the baseline BERT model, to our knowledge, there are currently no studies that sought to explain what tokens in the textual speech data the model draws attention from to perform its classifications on CPS dimensions. The lack of understanding to explain the inner workings of such black-box models poses a significant barrier to their adoption by practitioners, especially in the field of education \cite{cukurova2020}.

\subsection{Scope and research question}
In this study, we initiate this line of inquiry by discussing how tokens within different utterances of textual speech data contribute to the model's performance for CPS diagnosis. In particular, we investigate two research questions:
\begin{enumerate}
    \item[RQ1:] How are words with high contribution to classification on affective dimensions and well-performing classifications on social-cognitive dimensions associated with the contextual meaning of those classes?
    \item[RQ2:] What words does the model most frequently attend to when making its classifications?
\end{enumerate}
This study aims to provide insights into how a fine-tuned BERT base model leverages tokenised utterances to classify data associated with CPS classes. Through the identification of such words, the goal is to highlight the limitations inherent in using the opaque large language BERT model for CPS diagnosis, and to discuss the implications for its adoption by teachers.

\section{Method}

To enable this investigation, the conceptual framework used to code textual data would need to be as specific as possible (i.e., indicator-level coding at the utterance level) and provide some degree of differentiation in terms of the processes undertaken during CPS. For example, students' conversations when trying to solve a problem are situated differently from when they are following a set of instructions on how to solve a problem. Such distinctions would allow for more precise identification and interpretation of why specific tokens emerge and are used for classification. Hence, this study is based on a dataset comprising utterance-level codings of fine-grained indicators of CPS processes across three dimensions: problem solving subskills, scripting subskills and affective states (see \cite{wong_2025Rethinking} for CPS framework). A fine-tuned BERT model trained on such a dataset would be relevant in this study to examine the contribution of tokenised words to model classification.

Each dimension was treated as an independent single-label multi-class classification task, in which a BERT model was used to predict a single class from the social-cognitive dimension, and a separate BERT model predicted a single class from the affective dimension. Table \ref{tab:dataset} provides a breakdown of the different classes represented in the training and testing datasets, along with the weighted F1 score for each class by the BERT model considered in this study. Prior to classification, Named Entity Recognition (NER) with human correction was applied to the transcription dataset to annotate masks associated with names, web-based resources, locations, entertainment and devices that were mentioned during CPS. This was done to reduce noise in the dataset \cite{mohit_2014Named} and enable the BERT model to make classifications on meaningful tokenised words.

\begin{table}[h]
\caption{Dataset of classes used for training its corresponding weighted F1 scores from classification on testing data. The scores above 0.6 are indicated in bold face.}
\label{tab:dataset}
\small
\renewcommand{\arraystretch}{1.2} 
\setlength{\tabcolsep}{2.5pt} 
\centering
\resizebox{0.47\textwidth}{!}{
\begin{tabular}{c|cccccccccc|ccc}
\hline
Dimension     & \multicolumn{10}{c|}{Social-cognitive}                      & \multicolumn{3}{c}{Affective} \\
Class       & SS1 & SS2  & SS3 & SS4 & SS5 & SS6 & SS7 & SS8 & SC1  & SC2 & AS1      & AS2      & AS3     \\ \hline
Training & 172 & 978  & 279 & 21  & 75  & 101 & 18  & 41  & 1272 & 687 & 455      & 736      & 31      \\
Testing  & 43  & 245  & 70  & 6   & 19  & 25  & 4   & 10  & 318  & 172 & 114      & 184      & 8       \\  
F1 score & .185 & \textbf{.624}  & .299 & .286  & .452  & .069 & 0  & 0  & \textbf{.698} & \textbf{.624} & \textbf{.878 }     & \textbf{.914 }     & .400      \\ \hline      
\end{tabular}
}
\end{table}

From Table \ref{tab:dataset}, we observe that only three of the classes in the social-cognitive dimension had a weighted F1 score above $0.6$ (i.e., well-performing classifications): `\textit{Building shared understanding}' (SS2), `\textit{Using scripting}' (SC1), and `\textit{Regulating scripting}' (SC2). The remaining CPS classes appeared to be poorly classified, suggesting that the model may struggle to identify these classes effectively from the tokenised words. Hence, we focus our investigation of the tokenised words only on these three classes in the social-cognitive dimension. We also included the three affective states in the affective dimension in our investigation, `\textit{Neutral affective state}' (AS1), `\textit{Negative affective state}' (AS2), and `\textit{Positive affective state}' (AS3). 

\subsection{Approach for model explanation}
One approach to model interpretability in Explainable AI (XAI) is the post-hoc method, where explanations are determined after a model has been trained. Since this study seeks to understand the general model behaviour and elucidate the logic behind its classification, the SHapley Additive exPlanations (SHAP) \cite{Lundberg2017} approach was used to obtain Shapley values that are indicative of the feature importance of tokenised words across multiple utterances in the dataset. SHAP is model-agnostic and only requires the inputs and outputs to estimate the feature importance values (i.e., Shapley values). Its strengths and limitations have been widely studied in the field of XAI \cite{salih2024,Slack2020} and are applied extensively in various fields such as healthcare \cite{vimbi_2024Interpreting}, natural language processing \cite{bias_bulath} and education \cite{bulathwela2020predicting}. In particular, we performed a preliminary investigation using \texttt{shap.Explainer}\footnote{\url{https://shap.readthedocs.io/en/latest/generated/shap.Explainer.html}}, which automatically selects the most suitable algorithm (e.g., permutation, partition, tree, or linear) based on the given model and data type to approximate the Shapley values. The \texttt{PartitionExplainer}\footnote{\url{https://shap.readthedocs.io/en/latest/generated/shap.PartitionExplainer.html}} algorithm was selected and applied to compute Shapley values recursively through a hierarchy of tokenised words in the utterances. The advantage of this approach lies in its shorter computational runtime compared to other well-known algorithms (e.g., KernalSHAP).

The research questions focus on understanding how the tokenised words processed by the BERT model affected the classification performance. Hence, SHAP is applied to the testing dataset, as it contains unseen data used to evaluate the model's performance. Applying SHAP to the training dataset could result in misleading feature importance since the model may have overfitted to the training data. This may cause features that do not generalise beyond the data on which the model was trained to become identified.

\subsection{Analysis}
After applying SHAP, we obtained a list of Shapley values for each tokenised word from BERT. A positive Shapley value for a word indicates that it is pushing the model towards the classification (positive contribution). In contrast, a negative Shapley value indicates that the token is pushing the model away from the classification (negative contribution). Values that are close to zero have little or no contribution towards the model's classification. For each class, the Shapley values of a tokenised word were averaged across all the different utterances to obtain an aggregated Shapley value (\textit{avgSHAP}). This represented the overall feature importance of that word for the specified class. 

The analysis focused on the top 20 words with the highest overall feature importance for the classifications, as determined by the absolute \textit{avgSHAP} values. This allowed sufficient representation of the critical words to be considered, as this allowed the word with the smallest \textit{avgSHAP} value in the list to have an absolute value that was about half of the highest absolute \textit{avgSHAP} value in the list. These words, along with their respective \textit{avgSHAP} values, were obtained from the SHAP boxplot and consolidated into a table for comparison.

Among the top 20 words of each class, the top 10 words with positive \textit{avgSHAP} values were labelled in decreasing order of absolute value from P1 to P10. Similarly, the words with negative \textit{avgSHAP} values were also labelled in decreasing order of absolute value from N1 to N10. Using these labels, a layered heatmap is used to visualise the presence of these words across the classifications, with its vertical axis as the classes and horizontal axis as the top contributing words that appeared in at least $50\%$ of the classes (i.e., five or more times across all 10 social-cognitive classes and twice or more times across all three affective classes). For each class, positive \textit{avgSHAP} values with the highest absolute magnitude are represented in dark green, transitioning to light green as the magnitude decreases (i.e., P1 - P10). The negative \textit{avgSHAP} values with the highest absolute magnitude are represented in dark orange, with decreasing magnitudes shown in progressively lighter shades of orange (i.e., N1 - N10).

\section{Results}
\subsection{Words in well-performing classifications}
\subsubsection{Social-cognitive dimensions}
Here, Table \ref{tab:SS1SC1SC2} shows the top 20 tokenised words in the utterances that contributed to the predictions for SS2, SC1 and SC2 based on the \textit{avgSHAP} values.

\begin{table}[ht]
    \caption{Top 20 words that contribute towards the classification out of 805 other words based on \textit{avgSHAP} values from the SHAP bar plot of SS2, SC1 and SC2. The words are arranged in decreasing order of absolute magnitude of \textit{avgSHAP} value, with negative contributions coloured red.}
    \label{tab:SS1SC1SC2}
\centering
\resizebox{0.47\textwidth}{!}{
    \begin{tabular}{cc|cc|cc}
     \hline
 \multicolumn{2}{c|}{\textit{SS2: Building Shared Understanding}}&\multicolumn{2}{c|}{\textit{SC1: Using Scripting}}& \multicolumn{2}{c}{\textit{SC2: Regulating Scripting}}\\
 \hline
 Word& \textit{avgSHAP} value& Word& \textit{avgSHAP} value& Word&\textit{avgSHAP} value\\
     \hline
          compare&  0.81&  derive& 0.82
& \textcolor{red}{hypothesis}& \textcolor{red}{-0.57} 
\\
          therefore&  0.69&  hypothesis& 0.79
& \textcolor{red}{derive}& \textcolor{red}{-0.57} 
\\
          radius&  0.60&  topics& 0.71
& \textcolor{red}{compare}& \textcolor{red}{-0.55} 
\\
          points&  0.52&  relevant& 0.64
& \textcolor{red}{therefore}& \textcolor{red}{-0.50}
\\
          diagram&  0.52&  researcher& 0.62
& \textcolor{red}{cancer}& \textcolor{red}{-0.48}
\\
          angle&  0.47&  descriptive& 0.53
& \textcolor{red}{descriptive}& \textcolor{red}{-0.44} 
\\
          middle&  0.46&  solution& 0.53
& \textcolor{red}{researcher}& \textcolor{red}{-0.43} 
\\
          arc&  0.45&  contributing& 0.52
& \textcolor{red}{overall}& \textcolor{red}{-0.43 }
\\
  wheelbase& 0.41& contribute&0.48
& \textcolor{red}{relevant}& \textcolor{red}{-0.42}
\\
  curve& 0.40& support&0.45&  
\textcolor{red}{35}& \textcolor{red}{-0.42}
\\
 point& 0.38& collaborate& 0.45& \textcolor{red}{actual}&\textcolor{red}{-0.4}\\
 width& 0.37& cancer& 0.45& \textcolor{red}{key}&\textcolor{red}{-0.4}\\
 \textcolor{red}{exam}& \textcolor{red}{-0.37}& respond& 0.44& \textcolor{red}{wrong}&\textcolor{red}{-0.38}\\
 upwards& 0.37& chose& 0.43& \textcolor{red}{divided}&\textcolor{red}{-0.38}\\
 was& 0.35& \textcolor{red}{drive}& \textcolor{red}{-0.43}& \textcolor{red}{factor}&\textcolor{red}{-0.36}\\
 triangle& 0.35& strongly& 0.42& \textcolor{red}{solution}&\textcolor{red}{-0.35}\\
 simultaneous& 0.35& copied& 0.41& 02&0.35\\
 quarter& 0.32& actual& 0.41& \textcolor{red}{contributing}&\textcolor{red}{-0.33}\\
 circular& 0.32& achieve& 0.41& \textcolor{red}{gather}&\textcolor{red}{-0.32}\\
 key& 0.32& helps& 0.40& ten&0.32\\
  \hline
    \end{tabular}
    }
\end{table}
It was observed that there were no common words between SS2 and SC1. Since SS2 was focused on the problem solving dimension, whereas SC1 was on the scripting dimension, the BERT model was able to diagnose specific words that were distinct between these two classes. A further analysis was performed by determining the number of times these words appeared in the problem solving question and in the scripting instructions that were provided in the CPS task. It was observed that some of the words that contributed positively to the classification of `\textit{Building shared understanding}' were present in the problem solving task: \textit{radius} (6 times), \textit{diagram} (3 times), \textit{angle} (1 time), and \textit{wheelbase} (2 times). None of these words were present in the scripting instructions. On the contrary, the words that contributed positively to the classification of `\textit{Using scripting}' were used in the scripting instructions and not in the problem solving task: \textit{researcher} (1 time), \textit{contribute} (1 time) and \textit{support} (1 time). The word `\textit{solution}' appeared only once in the problem solving task but 5 times in the scripting instruction. These suggest that the BERT model was able to differentiate between problem solving and scripting terms that were used by students during CPS, making distinct associations of words that were contextually meaningful to these classes; `\textit{Building shared understanding}' would involve discussion and usage of terms in the problem solving question and `\textit{Using scripting}' would involve discussion on following through with given instructions in the CPS task.

It was also seen that the list of words for SC2 mostly had negative \textit{avgSHAP} values. These words mostly had positive \textit{avgSHAP} values for classifications in SS2 and SC1. The two words `02' and `10' do not have any meaningful association with SC2. Classifications for '\textit{Regulating scripting}' remain largely unclear and could be based on natural conversations that occurred among students, which may not have clear distinguishing words that allow for a meaningful explanation.

\subsubsection{Affective dimensions}
Table \ref{tab:afposnegneut} shows the top 20 tokenised words in the utterances that contributed to the classification of AS1, AS2 and AS3 based on the \textit{avgSHAP} values.

\begin{table}[ht]
    \caption{Top 20 words that contribute towards the classification out of 278 other words based on \textit{avgSHAP} values from the SHAP bar plot of AS1, AS2 and AS3. The words are arranged in decreasing order of absolute magnitude of \textit{avgSHAP} value, with negative contributions coloured red.}
    \label{tab:afposnegneut}
\centering
\resizebox{0.47\textwidth}{!}{
    \begin{tabular}{cc|cc|cc}
     \hline
 \multicolumn{2}{c|}{\textit{AS1: Neutral Affective State}}&\multicolumn{2}{c|}{\textit{AS2: Negative Affective State}}& \multicolumn{2}{c}{\textit{AS3: Positive Affective State}}\\
 \hline
 Word& \textit{avgSHAP} value& Word& \textit{avgSHAP} value& Word&\textit{avgSHAP} value\\
     \hline
          \textcolor{red}{huh}
&  \textcolor{red}{-0.77}&  huh& 0.78& \textcolor{red}{hate}& \textcolor{red}{-0.07}\\
          \textcolor{red}{wah}
&  \textcolor{red}{-0.75}&  wah& 0.76& congratulations& 0.07\\
          \textcolor{red}{uh}
&  \textcolor{red}{-0.61}&  uh& 0.62& great& 0.06\\
          \textcolor{red}{shit}
&  \textcolor{red}{-0.59}&  shit& 0.61& together& 0.06\\
          \textcolor{red}{nothing}
&  \textcolor{red}{-0.46}&  nothing& 0.50& \textcolor{red}{nothing}& \textcolor{red}{-0.05}\\
          \textcolor{red}{eh}
&  \textcolor{red}{-0.44}&  eh& 0.48& \textcolor{red}{stop}& \textcolor{red}{-0.05}\\
          \textcolor{red}{whatever}
&  \textcolor{red}{-0.44}&  whatever& 0.45& \textcolor{red}{confused}& \textcolor{red}{-0.05}\\
          \textcolor{red}{impossible}
&  \textcolor{red}{-0.42}&  impossible& 0.44& all& 0.05\\
  \textcolor{red}{hai}
& \textcolor{red}{-0.38}& hai&0.38& \textcolor{red}{confusing}& \textcolor{red}{-0.04}\\
  \textcolor{red}{ur}& \textcolor{red}{-0.35}& ur&0.37&  
\textcolor{red}{zero}& \textcolor{red}{-0.04}\\
 \textcolor{red}{ah}& \textcolor{red}{-0.33}& what& 0.35& \textcolor{red}{loss}&\textcolor{red}{-0.04}\\
 \textcolor{red}{what}& \textcolor{red}{-0.33}& oh& 0.35& supportive&0.03\\
 \textcolor{red}{oh}& \textcolor{red}{-0.32}& ah& 0.34& most&0.03\\
 \textcolor{red}{wal}& \textcolor{red}{-0.25}& hate
& 0.30& are&0.03\\
 \textcolor{red}{hate}& \textcolor{red}{-0.23}& wal& 0.25& \textcolor{red}{here}&\textcolor{red}{-0.03}\\
 \textcolor{red}{god}& \textcolor{red}{-0.23}& god& 0.25& \textcolor{red}{stupid}&\textcolor{red}{-0.03}\\
 \textcolor{red}{zero}& \textcolor{red}{-0.20}& zero& 0.25& \textcolor{red}{doomed}&\textcolor{red}{-0.03}\\
 \textcolor{red}{congratulations}& \textcolor{red}{-0.19}& confusing& 0.23& team&0.03\\
 \textcolor{red}{confusing}& \textcolor{red}{-0.19}& stop& 0.23& \textcolor{red}{hell}&\textcolor{red}{-0.03}\\
 \textcolor{red}{ao}& \textcolor{red}{-0.18}& confused& 0.23& \textcolor{red}{oh}&\textcolor{red}{-0.03}\\
  \hline
    \end{tabular}
    }
\end{table}

The list of words with positive \textit{avgSHAP} values for AS2 were closely associated with indicators of affective states: verbal cues to express bewilderment (i.e., `\textit{huh}', `\textit{wah}', `\textit{uh}', `\textit{eh}', `\textit{ur}', `\textit{what}', `\textit{oh}', `\textit{ah}'), vulgar expressions (i.e., `\textit{shit}', `\textit{god}'), unwillingness to engage (i.e., `\textit{nothing}', `\textit{whatever}', `\textit{stop}') and verbalising difficulty (i.e., `\textit{impossible}', `\textit{confusing}', `\textit{confused}', and `\textit{hai}' which denotes sighing). There was a coherent explanation of the list of words that contributed towards the classification of AS2. This provides confidence in the BERT model for the high classification performance of AS2 (i.e., having the highest weighted F1 score of 0.914 among all the classifications in the affective dimension).

In AS3, the only words among the top 20 words that had positive \textit{avgSHAP} values were `\textit{congratulations}', `\textit{great}', `\textit{together}', `\textit{all}', `\textit{supportive}', `\textit{most}', `\textit{are}', `\textit{team}'. Except for the word `\textit{are}', these words convey the notion of satisfaction and make references to the collective group. This is congruent with the indicator coded in the positive affective state (i.e., self-praise and complimenting or encouraging others). On the other hand, the words with negative \textit{avgSHAP} values were observed to have a close association with negative affective states and were in the list of word with positive \textit{avgSHAP} values in AS2 (e.g., `\textit{hate}', `\textit{confused}', \textit{stupid}', `\textit{doomed}', and `\textit{hell}'). Interestingly, in the list of words for AS3, both \textit{avgSHAP} values for the words that had positive contributions and negative contributions were all close to 0. This suggests that although the contributions of the words may be small (based on Shapley values), they were sufficient to influence the classification of AS3. The observed tokenised words adequately account for how the model diagnosed the positive affective state.

Similar to the classification on SC2, the list of words for the classification of AS1 all had negative \textit{avgSHAP} values. These words were largely the same list of words that contributed positively to the classification of AS2; with the exception of `\textit{congratulations}' that contributed positively to the classification of AS3, and two tokenised words `\textit{wal}' and `\textit{ao}' that were tokenised from `walao' to express surprise (used in Singapore colloquial language). Hence, the model may be interpreting that the neutral affective state was predominantly the opposite of the negative and positive affective state (i.e., words not associated with the positive or negative affective state are representative of the neutral affective state).

\subsection{Frequently occurring words across classes}
\begin{figure*}[!h]
    \centering
    \includegraphics[width=1\linewidth, trim=40 0 0 0, clip]{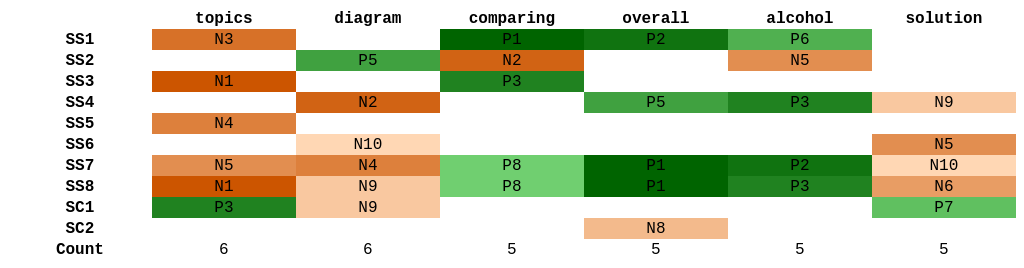}
    \caption{Top words occurring in at least 5 of the classes in the social-cognitive dimension with annotation of their relative contribution within the class based on its \textit{avgSHAP} values.}
    \label{fig:scheat}
\end{figure*}

\begin{figure*}[!h]
    \centering
    \includegraphics[width=1\linewidth, trim=20 0 0 0, clip]{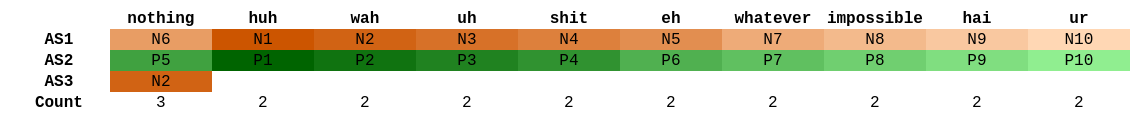}
    \caption{Top words occurring in at least 2 of the classes in the affective dimension with annotation of their relative contribution within the class based on its \textit{avgSHAP} values.}
    \label{fig:affectheat}
\end{figure*}

By analysing the list of words that occurred across all the social-cognitive classes and affective classes, we obtain the heatmaps in Figure \ref{fig:scheat} and Figure \ref{fig:affectheat} respectively.

We observe that the heatmap for affective classes was highly ordered compared to that of social-cognitive classes, in which the words had negative contributions to AS1 and AS3, and only had positive contributions to AS2. There was no such pattern observed among words in the social-cognitive classes. However, we observed that the words `\textit{comparing}', `\textit{overall}' and `\textit{alcohol}' were determined by the model to contribute positively to the classification of the following classes: `\textit{Sense-making}' (SS1), `\textit{Maintaining shared understanding}' (SS7), and `\textit{Evaluating the solution}' (SS8). Although the words `\textit{comparing}' and `\textit{overall}' would be relevant to these classes when students are trying to progress towards answering the question, these words were not useful to enable a clear distinction between these classes. This could explain why SS1 had a low F1 score (i.e., 0.185), and there were no classifications on SS7 and SS8. Compared to well-performing classifications, the word `\textit{diagram}' contributed positively to only SS2, and both the words `\textit{topics}' and `\textit{solution}' contributed positively to only SC1.

However, this raises the question of why the term `\textit{alcohol}' was identified by the model to contribute positively to the classification of problem solving classes. Further inspection of students' dialogue showed that when the problem solving question asked students to perform calculations on the width of the road to allow vehicles to drive on safely, they talked about safe driving in relation to the use of alcohol. As the BERT model is a large language model pre-trained on an extensive corpus of text data, the word `\textit{alcohol}' could have been associated with the word `driving' used in the problem-solving question. A semantic similarity between the term `\textit{alcoholic}' and `\textit{driving}' was performed using pre-trained word embeddings derived from the \texttt{en\_core\_web\_lg} model of the \texttt{spaCy} NLP library. The cosine similarity of $0.272$ between the two words indicates that they are conceptually distinct, yet exhibit some semantic relatedness. However, although the word `\textit{alcohol}' may appear contextually valid and reasonable in conversation among students, it may not be semantically meaningful for use in the classification of problem solving classes.

\section{Discussion}
We observe that well-detected classifications of the BERT model do not necessarily imply that there may be reasonable explanations for these classifications (as seen in the classification of SC2). The granularity of the indicators that are reasoned and distinguished by the coders may not be captured within the model architecture of the BERT base model. For example, differentiating between the use of scripting and regulating script use involves identifying whether the conversation related to scripting is situated in following through with the instructional activities or in organising the team around these activities. Such nuance is challenging for the BERT model to capture by text alone. One way to tackle this challenge is to leverage human-AI complementarity \cite{Cukurova2024}. Since we have observed that clear distinctions can be explained and made between the problem solving and scripting dimensions, the BERT model could first perform this differentiation in the automated diagnosis and leave the complex identification between SC1 and SC2 to teachers.

The increased number of classifications that the BERT model has to perform may result in the frequent use of particular tokenised words for classification. An ensemble model architecture that pays particular attention to smaller subsets of classes (especially for the problem solving dimension) might improve both model performance and explanation of the model's classification. In particular, such approaches could potentially reduce the risk of spurious words that would positively contribute to a given classification, but are not meaningful or relevant for use in the classification. Such words may potentially undermine the validity of the model and diminish the teachers' trust in relying on its classifications.

\subsection{Limitation and future work}
The use of XAI methods, specifically SHAP, helped uncover aspects of the inner workings of the BERT model's classifications. This approach provides valuable insights for model improvement and considerations that might remain obscured when evaluation is based solely on performance metrics (e.g., accuracy or weighted F1 scores). In this study, we applied the \texttt{PartitionExplainer} algorithm as a preliminary investigation on the use of SHAP to explain CPS diagnosis of a BERT model. Further studies could incorporate widely used other algorithms, such as the Kernel SHAP method \footnote{\url{https://shap.readthedocs.io/en/latest/generated/shap.KernelExplainer.html}} \cite{Lundberg2017}, to compare variations in feature importance across different methods and datasets within the CPS context.

Furthermore, additional investigations can be conducted to provide a more comprehensive explanation of CPS classifications in multimodal models. Multimodal models utilise different data modalities (e.g., textual speech and acoustic prosodic data from audio speech) that are encoded and processed through various layers (e.g., self-attention). Although such multimodal models may have the potential to improve model performance, providing explanations of the words used by the model to achieve this improved performance is very challenging. However, it would be very valuable for enhancing transparency and providing better directions for improving automated CPS diagnosis in the future.

Finally, it remains unclear how the provision of Shapley values may guide teachers in their understanding of how the model is making its classifications. Currently, the study has demonstrated how Shapley values can be applied to derive the words that contributed to the classifications. However, this information alone may not be sufficient to provide rigorous explanations to teachers. Future studies could consider approaches such as topic modelling \cite{shin_2025Explainable} to examine how the top contributing words are associated with the classes, and potentially provide an explainable evaluation of its classifications to teachers.

\begin{acknowledgments}
\small 
This work is funded by the European Commission's projects ``Teacher-AI Complementarity (TaiCo)" (Project ID: 101177268), ``Humane AI" (Grant No. 820437) and ``X5GON" (Grant No. 761758). The CPS framework is provided at \url{https://osf.io/u4nvb/?view\_only=3b02f769ea4746a0bbbae514ab1acb8e}.
\end{acknowledgments}

\section*{Declaration on Generative AI}
  The authors have not employed any Generative AI tools.


\end{document}